\documentclass{IOS-Book-Article}

\usepackage{mathptmx}
\usepackage{soul}\setuldepth{article}
\def\hb{\hbox to 11.5 cm{}}

\usepackage{xcolor}
\usepackage[colorinlistoftodos,prependcaption,textsize=tiny]{todonotes}

\usepackage[normalem]{ulem}
\usepackage{tcolorbox}
\usepackage{mathtools}
\usepackage{ulem}

\usepackage[numbers]{natbib}
\usepackage{multicol}
\PassOptionsToPackage{hidelinks}{hyperref}
\usepackage{graphicx}
\graphicspath{ {./images/} }

\usepackage{fancyvrb,fvextra}
\usepackage{float}
\floatstyle{ruled}
\newfloat{code}{thp}{lop}
\newfloat{result}{tbp}{lop}
\floatname{code}{Code}
\floatname{result}{Result}

\begin{document}

\pagestyle{headings}
\def\thepage{}
\begin{frontmatter}              

    \title{We, Vertiport 6, are temporarily closed: Interactional Ontological Methods for Changing the Destination} 

\markboth{}{Aug. 2023\hb}

\author[A]{\fnms{Seungwan} \snm{Woo}\orcid{0009-0006-7504-6750}},
\author[B]{\fnms{Jeongseok} \snm{Kim}\orcid{0000-0001-6431-7910}}
and
\author[A]{\fnms{Kangjin} \snm{Kim}\orcid{0000-0003-2912-6726}%
\thanks{Corresponding Author: Kangjin Kim, Email: kangjinkim@cdu.ac.kr}}

\runningauthor{S. Woo et al.}
\address[A]{Department of Drone Systems, Chodang University, Jeollanam-do, Korea}
\address[B]{AIX, SK Telecom, Seoul, Korea}

\begin{abstract}
This paper presents a continuation of the previous research on the interaction between a human traffic manager and the UATMS. In particular, we focus on the automation of the process of handling a vertiport outage, which was partially covered in the previous work. Once the manager reports that a vertiport is out of service, which means landings for all corresponding agents are prohibited, the air traffic system automates what it has to handle for this event. The entire process is simulated through knowledge representation and reasoning. Moreover, two distinct perspectives are respected for the human supervisor and the management system, and the related ontologies and rules address their interactions. We believe that applying non-monotonic reasoning can verify each step of the process and explain how the system works. After a short introduction with related works, this paper continues with problem formulation, primary solution, discussion, and conclusions.
\end{abstract}

\begin{keyword}
UAM, UATM, KRR, Answer Set Programming, Articulating Agent
\end{keyword}
\end{frontmatter}
\markboth{Aug. 2023\hb}{Aug. 2023\hb}

\section{Introduction}
Urban Air Mobility (UAM) is a new way to travel that is quickly becoming popular. It is regarded as a key solution to resolving traffic issues in crowded cities by utilizing urban airspace. Thanks to the concept of a vertical airport (or vertiport for short), small aircraft can take off and land around the compact buildings without runways.

As much as it is rapidly growing in popularity as a new mode of transportation, the pressure to standardize the UAM air traffic management system (or UATM for short) gets higher. At the same time, this becomes complicated in order to include the various needs of different stakeholders.

Introducing a new system into already well-developed environments sounds demanding. Particularly, the airspace should be shared with the existing aviation system. Possible conflicts between two systems are avoided by splitting the altitude to be used. Including the inherent challenges of low-altitude flights and high-density air traffic and resolving these issues, we plan to realize this new transportation system by 2035.

This research looks into the process of interactions between each vertiport and the UATM. We mainly want to discuss the role and procedure of vertiport managers and the UATM when an event occurs. Abstracting the traffic system through a graph-based one, we simply describe the system and provide a scenario that may happen frequently in a vertiport.

We have already taken our initial step in the previous paper. The main difference between the previous one and this is that while the previous one assumed certain corridors should be avoided to compute the route of the aircraft again, this considers a specific vertiport (which is the current destination) to be avoided. This naturally leads to changing its corresponding route.

\subsection{Related Works}

The papers \cite{Reiche2018, Garrow2021, KUAMConops10, Marzouk2022} give an in-depth examination of UAM technology, regulatory context, possible advantages, and obstacles.
Recently the article in \cite{FAAUAMConOps2} suggests a layered system for managing airspace for UAM vehicles, with different layers defined based on the type of vehicle, level of automation, and altitude.
The paper \cite{Schuchardt2023} discusses a variety of ATM projects in German Aerospace Center (DLR) initiatives.
The paper \cite{Kim2022} suggests a collision risk assessment model between a vehicle and obstacles.
The paper \cite{PintoNeto2023} discusses state-of-the-art Deep Learning (DL) solutions for ATM in depth.

However, none of previous attempts to avoid crashes based on ATM systems are ready to be widely adopted because of complex UATM systems, various stakeholders, mass agent activities, and possibly catastrophic events.
Considering multiple stakeholders, our initial work \cite{Kim2023Agent3C} was to detour agents' route, avoiding particular corridors. In that paper, nonmonotonic reasoning was utilized to illustrate route detours involving various UATMs.

In this regard, the paper \cite{BorregoDiaz2022} serves as a valuable guide for the explanation of complex systems---explainable AI (or XAI for short). To validate the complex system, their proposed epistemological model is based on knowledge representation and reasoning in particular. The paper \cite{Bourguin2021} examines a method for developing automatic classifiers capable of offering explanations based on an ontology. The paper \cite{Ozaki2020} briefly reviews five approaches based on association rule mining, formal concept analysis, inductive logic programming, computational learning theory, and neural networks, and provides how to build descriptive logic (DL) ontologies.

\subsection{Contributions}
As an illustration, we present a destination-changing scenario that may occur frequently in UAM environments. Then, we elaborate on our demonstration of nonmonotonic reasoning and explainability.

\subsection{Outlines}
In the following sections, we describe the problem formulation, provide the solution, continue discussions, and conclude this paper.

\section{Problem Formulation}
\begin{figure}
	\centering
	\includegraphics[width=0.85\linewidth]{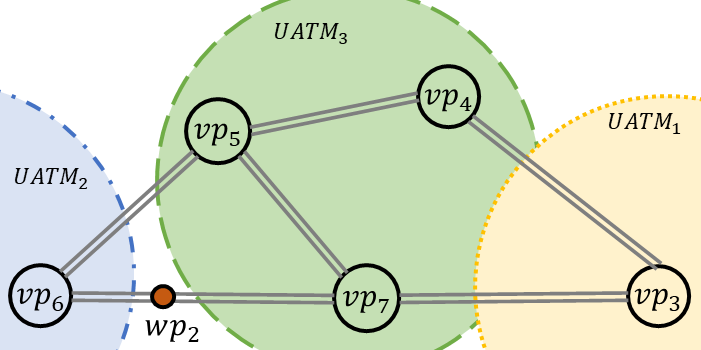}
	\caption{Partly sampled UATM Network composed of five vertiports ranging from $vp_3$ to $vp_7$, bidirectionally connected corridors between adjacent vertiports, three UATMs ranging from $UATM_1$ to $UATM_3$, and their coverage indicated by outside circles.}
	\label{fig:uatm_network_ep2}
\end{figure}

Examine the graph depicted in Fig.~\ref{fig:uatm_network_ep2}. Each vertex denotes a vertiport, while the edges denote corridors. Large circles represent UAM Air Traffic Management (UATM) systems covering these vertiports. These corridors transport agents between this network's vertiports. UATMs can communicate directly with agents within their coverage area. This study presumes that the `UATM Network' possesses a communication relaying protocol that enables UATMs to relay messages to one another. Upon the agent's departure from this UATM Network, direct communication becomes impossible.

Typically, agents begin a trip as follows: It begins in a vertiport, prepared for flight. This agent submits a takeoff request to the vertiport traffic manager. After authorization, the agent enters the airspace.

Prior to flight, the destination and route are configured. As it travels through corridors to the destination vertiport, the agent utilizes UATMs to update its velocity, GPS coordinates, and other data. UATMs can then monitor the agent while taking traffic into account.

When close, the agent requests landing from the vertiport traffic manager.

After confirmation, the traffic manager generally directs it to the vertipad. This traffic system functions independently. Human traffic managers are in charge of each vertical airport. They are responsible for monitoring traffic, vertiports, and legacy traffic systems. They enhance system adaptability.

\subsection{A Vertiport Closure Scenario}
\paragraph{\textbf{The human manager's perspective:}}
While a human manager in vertiport 6 (or $vp_6$ for short) monitored landing some agents, it found that there was a safety issue. It appears that approving landing requests from the following agents would be risky. Hence, he or she reported this status to the UATM (here, $UATM_2$). In particular, the $vp_6$ is temporarily closed, so it is required for all $vp_6$ heading agents to change their destination to $vp_5$, which is the candidate vertiport of the $vp_6$.

\paragraph{\textbf{The $UATM_2$'s perspective:}}
The $UATM_2$ received a report that the $vp_6$ is closed. Since all $vp_6$ heading agents should detour their route to the new destination, it should find all the related agents. While finding them, it also looks up other UATMs. Once it finds all, it checks the $vp_6$'s candidate vertiport. We assume that the candidate vertiports are carefully chosen, depending on previous records, including congestion, landing turnover ratio, and capacity. After getting the candidate vertiport and computing the route to the new destination, it sends to all these agents the replaced path, which has the newly appended route at the end. Once all these agents' paths are changed, the $UATM_2$ replies back to the $vp_6$ human manager.

\section{Solution}
In this section, the problem in Answer Set Programming is declared. We employ non-monotonic reasoning throughout the scenario modeling procedure for each small query. On the basis of the outcome and the incorporation of the queries, it is possible to determine if the provided mission is satisfiable. If the conditions are not met, the factors that contributed to their failure can be determined.

\subsection{Common Setting}
The whole approach begins with two common settings: one for the world and another for the data of the agents. Due to the page limitation, we included the Code~\ref{code:env_info} and Code~\ref{code:agent_info} as an appendix. It is necessary to execute these codes before any other code.

\subsection{Expressing in Ontology as the predicates}
We will provide several examples of natural language expressions that can be expressed in ontology as the predicates.

In Code~\ref{code:env_info}, predicates such as \texttt{uatm}, \texttt{vp}, \texttt{edge}, \texttt{cover}, \texttt{edge\_range}, \texttt{covered\_wp}, and \texttt{candidate\_vp} are related to the UATM Network.
In Code~\ref{code:agent_info}, predicates such as \texttt{loc}, \texttt{plan}, \texttt{source} and \texttt{target} are related to the agent.
Then, predicates such as \texttt{plan}, \texttt{new\_plan}, and \texttt{covered\_agent} are declared based on the necessity of the queries we will introduce.


The followings are a few examples for the ontology relationships and rules mentioned above.
\begin{itemize}
	\item "$UATM_1$ covers the vertiport 3" can be represented as \texttt{cover(1, 3)}.
	\item "A corridor from vertiport 7 to vertiport 6 has a range from 1 to 22" can be represented by \texttt{edge\_range(7, 6, 1..22)}.
	\item "Vertiport 6's candidate is vertiport 5" can be represented as \texttt{candidate\_vp(6, 5)}.
	\item "At step 1, agent 4 locates at waypoint 14 on the corridor between vertiport 7 and vertiport 6" can be represeted by \texttt{loc(4, 1, 7, 6, 14)}.
	\item "At step 1, agent 2 has a part of plan to move from vertiport 3 to vertiport 7" can be represented as \texttt{plan(2, 1, 3, 7)}.
	\item "$UATM_3$ covers a corridor from vertiport 7 to vertiport 6 within the range up to 12 from the beginning of that corridor's range" can be represented as the following rule: \texttt{cover\_wp(7, 6, 3, P) :- edge\_range(7, 6, P), P <= 12}. 
	
\end{itemize}

\subsection{Queries}
\paragraph{\textbf{Find all $vp_6$ heading agents within the UATM network:}}
In order to answer this query, there is an initial predicate \texttt{covered\_agent} in Code~\ref{code:query_01}. This rule is essential for determining whether each agent is covered by the UATM network. Additionally, it searches only agents whose target is $vp_6$.
Then, as shown in Result~\ref{res:query_01}, \texttt{covered\_by\_uatm2} and \texttt{covered\_by\_other} present five agents: 3, 5, 6, 1, and 2.

\begin{code}
	\begin{Verbatim}[breaklines,fontsize=\footnotesize]
covered_agent(A, TM) :- loc(A, T, U, V, WP), covered_wp(U, V, TM, WP), target(A, T, V1), V1 == 6.
covered_by_uatm2(A) :- covered_agent(A, 2).
covered_by_other(A) :- loc(A, T, U, V, _), covered_agent(A, TM), TM != 2.

trigger_query :- covered_agent(A, TM).
covered :- 1 <= #count{A: covered_by_uatm2(A); A:covered_by_other(A)}.
:- trigger_query, not covered.

#show loc/5.
#show covered_by_uatm2/1.
#show covered_by_other/1.
	\end{Verbatim}
	\caption{Find all $vp_6$ heading agents within the UATM network}
	\label{code:query_01}
\end{code}

\begin{result}
	\begin{Verbatim}[breaklines,fontsize=\footnotesize]
$ clingo env_info.lp agent_info.lp query01.lp
clingo version 5.6.2
Reading from env_info.lp ...
Solving...
Answer: 1
covered_by_other(3) covered_by_other(5) covered_by_other(6) covered_by_uatm2(1) covered_by_uatm2(2) loc(1,1,7,6,20) loc(2,1,7,6,18) loc(3,1,7,6,8) loc(4,1,7,6,14) loc(5,1,3,7,17) loc(6,1,7,6,3)
SATISFIABLE

Models       : 1
Calls        : 1
Time         : 0.012s (Solving: 0.00s 1st Model: 0.00s Unsat: 0.00s)
CPU Time     : 0.016s
	\end{Verbatim}
	\caption{All agents covered by UATM}
	\label{res:query_01}
\end{result}

\paragraph{\textbf{Change the destination of the agents we found:}}
Since we already found these agents from Code~\ref{code:query_01}, the next procedure for the $UATM_2$ has the followings:
\begin{itemize}
	\item it finds the candidate vertiport for the $vp_6$,
	\item it creates the new plans for the relayed agents: Here, relayed agents include agents covered by other UATMs as well as agents covered by $UATM_2$,
	\item it appends the new plan to the end of the existing plan for these agents,
	\item it sends target change requests to other UATMs, and
	\item once the plan is ready and target change requests are received, UATMs do the target change for these agents.
\end{itemize}

The $vp_6$ human manager, then, is known by the \texttt{target\_change} as shown in the Result~\ref{res:query_02}.

\begin{code}
	\begin{Verbatim}[breaklines,fontsize=\footnotesize]
relayed(A) :- covered_by_uatm2(A).
relayed(A) :- covered_by_other(A).

new_plan(A, T+1, V, V1) :- plan(A, T, U, V), target(A, T, V), V == 6, relayed(A), candidate_vp(V, V1), step(T+1), not new_plan(A, T, V, V1).
target_change_request(A, T) :- relayed(A), new_plan(A, T, V, V1).

plan(A, T+1, V, V1) :- plan(A, T, U, V), target(A, T, V), new_plan(A, T+1, V, V1), step(T+1).
plan(A, T+1, U, V) :- plan(A, T, U, V), step(T+1).

target_change(A, T) :- plan(A, T, U, V), new_plan(A, T, U, V), target_change_request(A, T).
:- not target_change(A, T), new_plan(A, T, U, V), target_change_request(A, T).

#show relayed/1.
#show new_plan/4.
#show target_change_request/2.
#show target_change/2.
	\end{Verbatim}
	\caption{Change the destination of the found agents}
	\label{code:query_02}
\end{code}

\begin{result}
	\begin{Verbatim}[breaklines,fontsize=\footnotesize]
$ clingo env_info.lp agent_info.lp query01.lp query02.lp
clingo version 5.6.2
Reading from env_info.lp ...
Solving...
Answer: 1
relayed(6) relayed(5) relayed(3) relayed(2) relayed(1) new_plan(1,2,6,5) new_plan(2,2,6,5) new_plan(3,2,6,5) new_plan(5,2,6,5) new_plan(6,2,6,5) target_change_request(1,2) target_change_request(2,2) target_change_request(3,2) target_change_request(5,2) target_change_request(6,2) target_change(1,2) target_change(2,2) target_change(3,2) target_change(5,2) target_change(6,2)
SATISFIABLE

Models       : 1
Calls        : 1
Time         : 0.007s (Solving: 0.00s 1st Model: 0.00s Unsat: 0.00s)
CPU Time     : 0.000s
	\end{Verbatim}
	\caption{Agents' targets are changed}
	\label{res:query_02}
\end{result}

\paragraph{\textbf{Handle a landing request from agent 4:}}
Consider the $vp_6$ human manager received a landing request from agent 4 shortly after he or she informed the \texttt{target\_change}. The agent 4 was in waypoint 2 (or $wp_2$ for short) in Fig.~\ref{fig:uatm_network_ep2} when $UATM_2$ searched all the $vp_6$ heading agents. We remark that even though every inch in each corridor is guaranteed to be within the coverage of the UATM network, relaying the message for each agent may take some time, and some of the agents may send the landing request before their targets are changed.

In the $vp_6$ manager's perspective, the landing request would not be welcome. However, after checking that its target hasn't been updated, he or she can just report to the $UATM_2$ that the $vp_6$ is closed again.

In the $UATM_2$'s perspective, since agent 4 is in the coverage of $UATM_2$ now, handling the \texttt{landing\_request} in Code~\ref{code:query_03} is not much different from what we have seen in Code~\ref{code:query_01} and Code~\ref{code:query_02}. The \texttt{target\_change(4, 3)} in Result~\ref{res:query_03} explains that agent 4 has a new target at step 3.

\begin{code}
	\begin{Verbatim}[breaklines,fontsize=\footnotesize]
vp6_heading_agent_number(N) :- N = #count{A:target(A, T, V), V==6}.

loc(4, 2, 7, 6, 17).
landing_request(A, T+1, V) :- not target(A, T+1, _), target(A, T, V), loc(A, T+1, U, V, WP), V == 6, covered_wp(U, V, TM, WP).

new_plan(A, T+1, V, V1) :- plan(A, T, U, V), landing_request(A, T, V), candidate_vp(V, V1), step(T+1), not new_plan(A, T, V, V1).
plan(A, T+1, V, V1) :- plan(A, T, U, V), landing_request(A, T, V), new_plan(A, T+1, V, V1), step(T+1).

target_change_request(A, T+1) :- landing_request(A, T, V), new_plan(A, T+1, V, V1).
plan(A, T+1, V, V1) :- plan(A, T, U, V), landing_request(A, T, V), new_plan(A, T+1, V, V1), target_change_request(A, T+1), step(T+1).
:- not target_change(A, T+1), landing_request(A, T, V), step(T+1).

#show vp6_heading_agent_number/1.
#show target_change_request/2.
#show target_change/2.
#show landing_request/3.
	\end{Verbatim}
	\caption{Handle the landing request from agent 4}
	\label{code:query_03}
\end{code}

\begin{result}
	\begin{Verbatim}[breaklines,fontsize=\footnotesize]
$ clingo env_info.lp agent_info.lp query01.lp query02.lp query03.lp
clingo version 5.6.2
Reading from env_info.lp ...
Solving...
Answer: 1
landing_request(4,2,6) new_plan(4,3,6,5) target_change_request(4,3) relayed(6) relayed(5) relayed(3) relayed(2) relayed(1) new_plan(1,2,6,5) new_plan(2,2,6,5) new_plan(3,2,6,5) new_plan(5,2,6,5) new_plan(6,2,6,5) target_change_request(6,2) target_change_request(5,2) target_change_request(3,2) target_change_request(2,2) target_change_request(1,2) target_change(4,3) target_change(1,2) target_change(2,2) target_change(3,2) target_change(5,2) target_change(6,2) vp6_heading_agent_number(6)
SATISFIABLE

Models       : 1
Calls        : 1
Time         : 0.009s (Solving: 0.00s 1st Model: 0.00s Unsat: 0.00s)
CPU Time     : 0.000s
	\end{Verbatim}
	\caption{Agent 4 has a new target.}
\label{res:query_03}
\end{result}

\section{Discussion}
We here discuss current progress, limitations, and future directions.

\subsection{Nonmonotonicity}
Through the entire program, the initial background knowledge given is that vertiport 6 is temporarily closed. Hence, this is a given fact. In order to illustrate the nature of nonmonotony, agents' locations can vary. However, due to the unified demonstration of the codes and the result, the locations of each agent are not selectable while running the system model. Since all the corresponding rules in the queries we covered have a general declaration, we believe that this can be easily generalized by changing their locations through choice rules.

\subsection{Explainability}
The query asking whether '\textbf{Changing the destination of the found agents}' is successful is represented as the predicate \texttt{target\_change} and its supporting rule just followed by the predicate. 
The answer shows that \texttt{target\_change(A, T)} is true when it is satisfiable.
In this query, the validation of the explanation is checked by the combination of the predicate, \texttt{target\_change}, which is regarded as a fact when the body of the rule is true, and the safe rule, which ensures the fact's consistency.
Assuming that all the derived rules and relationships lead to the body of the rule as true, the logical consequence makes the predicate \texttt{target\_change} by also being connected, and this justifies the answer to the query.
We note that the explanation is somewhat abstract depending on the ontological features, and the deeper explanation is desirable.

\subsection{Future Directions}
The research related in the explanation of the complex system is in early stage yet. More scenarios are necessary. Agent movement should be considered more carefully, and the scenario can be reactively simulated with proper foundations. We hope a comprehensive air traffic management system is deliverable in the near future.

\section{Conclusions}
We have described a scenario involving UATMs. Through knowledge representation and reasoning, the scenario was simulated by expressing the predicates in ontology. Applying the Answer Set Programming framework, queries and their responses were examined from two distinct perspectives: those of a human manager and the UATM system. The discussion then turned to nonmonotonicity and explainability.

\section*{Acknowledgments}
This work is supported by the Korea Agency for Infrastructure Technology Advancement(KAIA) grant funded by the Ministry of Land, Infrastructure and
Transport (Grant RS-2022-00143965).


\bibliographystyle{unsrtnat}
\bibliography{references}
\begin{appendix}
	\section{Common Settings}
	\subsection{Information about the Environment}
\begin{code}
	\begin{Verbatim}[breaklines,fontsize=\footnotesize]
uatm(1..3). agent(1..20). vp(1..7).

% edge(VP_u, VP_v) :
%    there is a corridor from VP_u to VP_v
edge(3, 7). edge(7, 3). edge(3, 4). edge(4, 3). edge(7, 6). edge(6, 7).
edge(6, 5). edge(5, 6). edge(5, 4). edge(4, 5). edge(7, 5). edge(5, 7).

% cover(UATM_i, VP_u) :
%    UATM_i coveres VP_u
cover(1, 3).
cover(3, 7). cover(3, 4). cover(3, 5).
cover(2, 6).

% edge_range(VP_i, VP_j, P) :
%    corridor from VP_i to VP_j has range P.
edge_range(3, 7, 1..20). edge_range(7, 3, 1..20).
edge_range(7, 6, 1..22). edge_range(6, 7, 1..22).
edge_range(3, 4, 1..20). edge_range(4, 3, 1..20).
edge_range(7, 5, 1..16). edge_range(5, 7, 1..16).
edge_range(6, 5, 1..16). edge_range(5, 6, 1..16).
edge_range(5, 4, 1..18). edge_range(4, 5, 1..18).

% covered_wp(VP_u, VP_v, UATM_i, P) :
%    UATM_i covers a corridor from VP_u to VP _v within the range P.
covered_wp(3, 7, 1, P) :- edge_range(3, 7, P), P <= 12.
covered_wp(7, 3, 1, P) :- edge_range(7, 3, P), P >= 8.
covered_wp(3, 4, 1, P) :- edge_range(3, 4, P), P <= 14.
covered_wp(4, 3, 1, P) :- edge_range(4, 3, P), P >= 6.
covered_wp(7, 6, 3, P) :- edge_range(7, 6, P), P <= 12.
covered_wp(6, 7, 3, P) :- edge_range(6, 7, P), P >= 10.
covered_wp(7, 5, 3, P) :- edge_range(7, 5, P).
covered_wp(5, 7, 3, P) :- edge_range(5, 7, P).
covered_wp(5, 6, 3, P) :- edge_range(5, 6, P), P <= 6.
covered_wp(6, 5, 3, P) :- edge_range(6, 5, P), P >= 10.
covered_wp(5, 4, 3, P) :- edge_range(5, 4, P).
covered_wp(4, 5, 3, P) :- edge_range(4, 5, P).
covered_wp(7, 3, 3, P) :- edge_range(7, 3, P), P <= 8.
covered_wp(3, 7, 3, P) :- edge_range(3, 7, P), P >= 12.
covered_wp(4, 3, 3, P) :- edge_range(4, 3, P), P <= 6.
covered_wp(3, 4, 3, P) :- edge_range(3, 4, P), P >= 14.
covered_wp(6, 7, 2, P) :- edge_range(6, 7, P), P <= 5.
covered_wp(7, 6, 2, P) :- edge_range(7, 6, P), P >= 17.
covered_wp(6, 5, 2, P) :- edge_range(6, 5, P), P <= 7.
covered_wp(5, 6, 2, P) :- edge_range(5, 6, P), P >= 9.

% candidate_vp(VP_i, VP_j) :
%    VP_i's candidate vp is VP_j.
candidate_vp(6, 5). candidate_vp(5, 4). candidate_vp(7, 5).
candidate_vp(4, 3). candidate_vp(3, 7).

step(1..3).
	\end{Verbatim}
	\caption{Information about the Environment}
	\label{code:env_info}
\end{code}
\newpage
	\subsection{Information about the Agents}
\begin{code}
	\begin{Verbatim}[breaklines,fontsize=\footnotesize]
% loc(AGENT_a, STEP_t, VP_i, VP_j, WP_p) :
%    At STEP_t, AGENT_a locates at WP_p on the corridor between VP_i and VP_j.
loc(1, 1, 7, 6, 20).
loc(2, 1, 7, 6, 18).
loc(3, 1, 7, 6, 8).
loc(4, 1, 7, 6, 14).
loc(5, 1, 3, 7, 17).
loc(6, 1, 7, 6, 3).

% plan(AGENT_a, STEP_t, VP_i, VP_j) :
%    At STEP_t, AGENT_a has a part of plan to move from VP_i to VP_j.
plan(1, 1, 3, 7). plan(1, 1, 7, 6).
plan(2, 1, 3, 7). plan(2, 1, 7, 6).
plan(3, 1, 3, 7). plan(3, 1, 7, 6).
plan(4, 1, 4, 3). plan(4, 1, 3, 7). plan(4, 1, 7, 6).
plan(5, 1, 4, 3). plan(5, 1, 3, 7). plan(5, 1, 7, 6).
plan(6, 1, 7, 6).

source(A, 1, U) :- agent(A), plan(A, 1, U, V), not plan(A, 1, _, U).
target(A, 1, V) :- agent(A), plan(A, 1, U, V), not plan(A, 1, V, _).	
	\end{Verbatim}
	\caption{Information about the Agents}
	\label{code:agent_info}
\end{code}

\end{appendix}
\end{document}